# Simplified Gating in Long Short-term Memory (LSTM) Recurrent Neural Networks

Yuzhen Lu and Fathi M. Salem
*Circuits, Systems, and Neural Networks (CSANN) Lab*
*Department of Biosystems and Agricultural Engineering || Department of Electrical and Computer Engineering*
*Michigan State University*
*East Lansing, Michigan 48824, USA*

*Abstract* – The standard LSTM recurrent neural networks while very powerful in long-range dependency sequence applications have highly complex structure and relatively large (adaptive) parameters. In this work, we present empirical comparison between the standard LSTM recurrent neural network architecture and three new parameter-reduced variants obtained by eliminating combinations of the input signal, bias, and hidden unit signals from individual gating signals. The experiments on two sequence datasets show that the three new variants, called simply as LSTM1, LSTM2, and LSTM3, can achieve comparable performance to the standard LSTM model with less (adaptive) parameters.

## I. INTRODUCTION

Recurrent neural networks (RNN) have recently shown great promise in tackling various sequence modeling tasks in machine learning, such as automatic speech recognition [1-2], language translation [3-4], and generation of language descriptions for images [5-6]. Simple RNNs, however, are difficult to train using the stochastic gradient decent and have been reported to exhibit the so-called "vanishing" gradient and/or "exploding" gradient phenomena [7-8]. This has limited the ability of simple RNN to learn sequences with relatively long dependencies.

To address this limitation, researchers have developed a number of techniques in network architectures and optimization algorithms [9-11], among which the most successful in applications is the Long Short-term Memory (LSTM) units in RNN [9, 12]. A LSTM unit utilizes a "memory" cell that may maintain its state value over a long time, and a gating mechanism that contains three non-linear gates, namely, an *input*, an *output* and a *forget* gate. The gates' intended role is to regulate the flow of signals into and out of the cell, in order to be effective in regulating long-range dependencies and achieve successful RNN training. Since the inception of the LSTM unit, many modifications have been introduced to improve performance. Gers et al. [13] have introduced "peephole" connections to the LSTM unit that connects the memory cell to the gates so as to infer precise timing of the outputs. Sak et al. [14-15] introduced two recurrent and non-recurrent projection layers between the LSTM units layer and the output layer, which resulted in significantly improved performance in a large vocabulary speech recognition task.

Adding more components in the LSTM units architecture, however, may complicate the learning process and hinder understanding of the role of individual components. Recently, researchers proposed a number of simplified variants of the LSTM-based RNN. Cho et al. [3] proposed a two-gate related architecture, called Gated Recurrent Unit (GRU) RNN, in which the *input*, *forget*, and *output* gates are replaced by an *update* gate and a *reset* gate. Chung et al. [16] presented performance comparisons between LSTM and GRU RNNs, and observed that the latter performed comparably or even exceeded the former on the specific dataset used. These conclusions, however, still are being further evaluated using more experiments and over broader datasets. In exploring eight architectural variants of the LSTM RNN, Greff et al. [17] found that coupling the *input* and *forget* gates, as in the GRU model, and removing peephole connections, did not significantly impair performance. Furthermore, they report that the *forget* gate and the *output* activation are critical components. These findings were corroborated by the work of Jozefowicz et al. [18] who evaluated an extensive architectural designs of ten thousand different RNNs. In [18], the authors observed that the *output* gate was the least important compared to the *input* and *forget* gates, and suggested adding a bias of 1 to the *forget* gate to improve the performance of the LSTM RNN. Zhou et al. [19] proposed a Minimal Gate Unit (MGU), which has a minimum of one gate, namely, the *forget* gate architecture, created by merging the *update* and *reset* gates in the GRU model. Through evaluations on four different sequence data, the authors found that an RNN with the fewer parameters MGU model was at par with the GRU model in terms of (testing) accuracy. The authors, however, did not explicitly perform comparisons against the standard LSTM RNN. Recently, Salem [20] introduced a simple approach to simplifying the standard LSTM model focusing only on the gating signal generation. The gating signals can be used as general control signals to be specified by minimizing the loss function/criterion. Specifically, all three gating equations were retained but their parameters were reduced by eliminating one or more of the signals driving the gates. For simplicity, we shall call these three variants, LSTM1, LSTM2, and LSTM3 and will be detailed in section III below.

The paper presents a comparative evaluation of the standard LSTM RNN model with three new LSTM model variants. The evaluation and test results have been demonstrated on two public datasets which reveal that the LSTM model variants are comparable to the standard LSTM

RNN model in testing accuracy performance. We remark that these are initial tests and further evaluations and comparisons need to be conducted among the standard LSTM RNN and the three LSTM variants.

The remainder of the paper is organized as follows. Section II specifies the standard LSTM RNN architecture with its three gating signals. Section III describes the three LSTM variants, called LSTM1, LSTM2, and LSTM3, respectively. Section IV presents the experiments considered in this study. Section V details the comparative performance results. Finally, section VI summarizes the main conclusions.

## II. THE RNN LSTM ARCHITECTURE

The LSTM architecture considered here is similar to that in Graves et al. [2, 16-19] but without peep-hole connections. It is referred to as the standard LSTM architecture and will be used for comparison with its simplified LSTM variants [20].

The (dynamic) equations for the LSTM memory blocks are given as follows:

$$i_t = \sigma(U_i h_{t-1} + W_i x_t + b_i) \quad (1)$$
$$f_t = \sigma(U_f h_{t-1} + W_f x_t + b_f) \quad (2)$$
$$o_t = \sigma(U_o h_{t-1} + W_o x_t + b_o) \quad (3)$$
$$c_t = f_t * c_{t-1} + i_t * \tanh(U_c h_{t-1} + W_c x_t + b_c) \quad (4)$$
$$h_t = o_t * \tanh(c_t) \quad (5)$$

In these equations, the ($n$-d vectors) $i_t$, $f_t$ and $o_t$ are the *input* gate, *forget* gate, *output* gate at time $t$, eqns (1)-(3). Note that these gate signals include the logistic nonlinearity, $\sigma$, and thus their signals ranges is between 0 and 1. The $n$-d cell state vector, $c_t$, and its $n$-d activation hidden unit, $h_t$, at the current time $t$, are in eqns (4)-(5). The input vector, $x_t$, is an $m$-d vector, tanh is the hyperbolic tangent function, and * in eqns (4)-(5), denotes a point-wise (Hadamard) multiplication operator. Note that the gates, cell and activation all have the same dimension ($n$). The parameters of the LSTM model are the matrices ($U_*$, $W_*$) and biases ($b_*$) in eqns (1)-(5). The total number of parameters (i.e., the number of all the elements in $W_*$, $U_*$ and $b_*$), say $N$, for the standard LSTM, can be calculated to be

$$N = 4 \times (m \times n + n^2 + n) \quad (6)$$

where, again, $m$ is the input dimension, and $n$ is the cell dimension. This constitutes a four-fold increase in parameters in comparison to the simple RNN [16-20].

## III. THE RNN LSTM VARIANTS

While the LSTM model has demonstrated impressive performance in applications involving sequence-to-sequence relationships, a criticism of the standard LSTM resides in its relatively complex model structure with 3 gating signals and the number of its relatively large number of parameters [see eqn (6)]. The gates in fact replicate the parameters in the cell. It is observed that the gates serve as control signals and the forms in eqns (1)-(3) are redundant [20]. Here, three simplifications to the standard LSTM result in three LSTM variants we refer to them here as simply, LSTM1, LSTM2, and LSTM3. There variants are obtained by removing signals, and associated parameters in the gating eqns (1)-(3). For uniformity and simplicity, we apply the changes to all the gates identically:

*1) The LSTM1 model: No Input Signal*

Here the input signal and its associated parameter matrix are removed from the gating signals (1)-(3). We thus obtain the new gating equations:

$$i_t = \sigma(U_i h_{t-1} + b_i) \quad (7)$$
$$f_t = \sigma(U_f h_{t-1} + b_f) \quad (8)$$
$$o_t = \sigma(U_o h_{t-1} + b_o) \quad (9)$$

*2) The LSTM2 model: No Input Signal and No Bias*

The gating signals contain only the hidden activation unit in all three gates, identically.

$$i_t = \sigma(U_i h_{t-1}) \quad (10)$$
$$f_t = \sigma(U_f h_{t-1}) \quad (11)$$
$$o_t = \sigma(U_o h_{t-1}) \quad (12)$$

*3) The LSTM3 model: No Input Signal and No Hidden Unit Signal*

The gating signals contain only the bias term. Note that, as the bias is adaptive during training, it will include information about the state via the backpropagation learning algorithms or the co-state [24].

$$i_t = \sigma(b_i) \quad (13)$$
$$f_t = \sigma(b_f) \quad (14)$$
$$o_t = \sigma(b_o) \quad (15)$$

Compared to the standard LSTM, it can be seen that the three variants results in $3mn$, $3(mn+n)$ and $3(mn+n^2)$ fewer parameters, respectively, and consequently, reducing the computational expense.

## IV. EXPERIMENTS

The effectiveness of the three proposed variants were evaluated using two public datasets, MNIST and IMDB. The focus here is to demonstrate the comparative performance of the standard LSTM RNN and the variants rather than to achieve state-of-the-art results. Only the standard LSTM RNN

[2, 16-19] was used as a base-line model and compared with its three variants.

*A. Experiments on the MNIST dataset:*

This dataset contains 60,000 training set and 10,000 testing set of handwritten images of the digits (0-9). The training set contains the labelled class of the image available for training. Each image has the size of 28×28 pixels. The image data were pre-processed to have zero mean and unit variance. As in the work of Zhou et al. [19], the dataset was organized in two manners to be the input of an LSTM-based network. The first was to reshape each image as a one-dimensional vector with pixels scanned row by row, from the top left corner to the bottom right corner. This results in a long sequence input of length 784. The second requires no image reshaping but treated each row of an image as a vector input, thus giving a much shorter input sequence of length 28. The two types of data organization were referred to as pixel-wise and row-wise sequence inputs, respectively. It is noted that the pixel-wise sequence is more time consuming in training.

In the two training tasks, 100 hidden units and 100 training epochs were used for the pixel-wise sequencing input, while 50 hidden units and 200 training epochs were used for the row-wise sequencing input. Other network settings were kept the same throughout, including the batch size set to 32, RMSprop optimizer, cross-entropy loss, dynamic learning rate ($\eta$) and early stopping strategies. In particular, for the learning rate, it was set to be an exponential function of training loss to speed up training, specifically, $\eta = \eta_0 \times \exp(C)$, where $\eta_0$ is a constant coefficient, and $C$ is the training loss. For the pixel-wise sequence, two learning rate coefficients $\eta_0$=1e-3 and 1e-4 were considered as it takes relatively long time to train, while for the row-wise sequence, four $\eta_0$ values of 1e-2, 1e-3, 1e-4 and 1e-5 were considered. The dynamic learning rate is thus directly related to the training performance. At the initial stage, the training loss is typically large, thus resulting in a large learning rate ($\eta$), which in turn increases the stepping of the gradient further from the present parameter location. The learning rate decreases only as the loss functions decreases towards lower loss level and eventually towards an acceptable minima in the parameter space. Thus was found to achieve faster convergence to an acceptable solution. For the early stopping criterion, the training process would be terminated if there was no improvement on the test data over consecutive epochs, in our case we chose 25 epochs.

*B. Experiments on the IMDB dataset:*

This dataset consists of 50,000 movie reviews from IMDB, which have been labelled into two classes according to (the reviews) sentiment, positive or negative. Both training and test sets contain 25,000 reviews. These reviews are encoded as a sequence of word indices based on the overall frequency in the dataset. The maximum sequence length was set to 80 among the top 20,000 most common words (longer sequences were truncated while shorter ones were zero-padded at the end). Referring to an example in the Library Keras [21], an embedding layer with the output dimension of 128 was added as an input to the LSTM layer that contained 128 hidden units. The dropout technique [22] was implemented to randomly zero 20% of signals in the embedding layer and 20% of rows in the weight matrices (i.e., $U$ and $W$) in the LSTM layer. The model was trained for 100 epochs. Other settings remained the same as those in the MNIST data.

Training LSTMs for the two datasets were implemented by using the **Keras** package in conjunction with the **Theano** library (the implementation code and results are available at: https://github.com/jingweimo/Modified-LSTM).

## V. RESULTS AND DISCUSSION

*A. The MNIST dataset:*

Table I summarizes the accuracies on the test dataset for the pixel-wise sequence. At $\eta_0$=1e-3, the standard LSTM produced the highest accuracy, while at $\eta_0$=1e-4, both LSTM1 and LSTM2 achieved accuracies slightly higher than that by the standard LSTM. LSTM3 performed the worst in both cases.

TABLE I
THE BEST ACCURACIES OF DIFFERENT LSTM NETWORKS ON THE TEST SET AND CORRESPONDING PARAMETER SIZES OF THE LSTM LAYERS

| LSTMs | Learning coefficient ($\eta_0$) | | #Params |
|---|---|---|---|
| | 1e-3 | 1e-4 | |
| Standard | 0.9857 | 0.9727 | 40,800 |
| LSTM1 | 0.9609 | 0.9799 | 40,500 |
| LSTM2 | 0.7519 | 0.9745 | 40,200 |
| LSTM3 | 0.4239 | 0.5696 | 10,500 |

Examining the training curves revealed the importance of $\eta_0$ and the different responses of the LSTMs. As shown in Fig. 1, the standard LSTM performed well in the two cases; while LSTM1 and LSTM2 performed similarly poorly at $\eta_0$=1e-3, where both suffered serious fluctuations at beginning and dramatically lowered accuracies at the end. However, decreasing $\eta_0$ to 1e-4 circumvented the problem fairly well for LSTM1 and LSTM2. For LSTM3, both $\eta_0$=1e-3 and 1e-4 could not achieve successful training because of the fluctuation issue, suggesting that $\eta_0$ should be decreased further. As shown in Fig. 2 where 200 training epochs were executed, choosing $\eta_0$=1e-5 provided a steadily increasing accuracy with the highest test accuracy of 0.7404. Despite the accuracy that was still lower than other LSTMs variants. It is expected that the LSTM3 would achieve higher accuracies if longer training time was allowed. In essence LSTM3 has the lowest parameters but needs more training execution epochs to improve (testing) accuracy.

The fluctuation phenomenon observed above is a typical issue caused by a large learning rate, and is likely due to numerical instability where the (stochastic) gradient can no longer be approximated, and it can readily be resolved by decreasing the learning coefficient-- however, at the price of slowing down training. From the results, the standard LSTM seemed more resistant to fluctuations in modeling long-sequence data than the three variants, more likely due to the suitability of the learning rate coefficient. LSTM3 was the most susceptible to the fluctuation issue, however, and it requires a lower coefficient. Its optimal coefficient in this study appears to be between $\eta_0$=1e-5 and $\eta_0$=1e-4. Thus, further investigation may lead to benefits of using the LSTM3 to reap the benefit of its dramatically reduced model parameters (see Table I).

Overall, these findings have showed that the three LSTM variants were capable of handling a long-range dependencies sequence comparable to the standard LSTM. Due attention should be paid to tuning the learning rate to achieve higher accuracies.

TABLE II
THE BEST ACCURACIES OF DIFFERENT LSTM NETWORKS VARIANTS ON THE TEST SET AND THEIR CORRESPONDING PARAMETER SIZES

| LSTMs | Learning rate coefficient ($\eta_0$) | | | | # Params |
|---|---|---|---|---|---|
| | 1e-2 | 1e-3 | 1e-4 | 1e-5 | |
| Standard | 0.9506 | 0.9816 | 0.9756 | 0.9555 | 15,800 |
| LSTM1 | 0.9820 | 0.9821 | 0.9730 | 0.9580 | 11,600 |
| LSTM2 | 0.9828 | 0.9799 | 0.9723 | 0.9580 | 11,450 |
| LSTM3 | 0.9691 | 0.9762 | 0.9700 | 0.9399 | 4,100 |

Among the four $\eta_0$ values, the $\eta_0$=1e-3 gave the best results for all the LSTMs except LSTM2 that performed the best at $\eta_0$=1e-2. Fig. 3 shows the corresponding learning curves at $\eta_0$=1e-3. All the LSTMs exhibited similar training pattern profiles, which demonstrated the efficacy of the three LSTM variants.

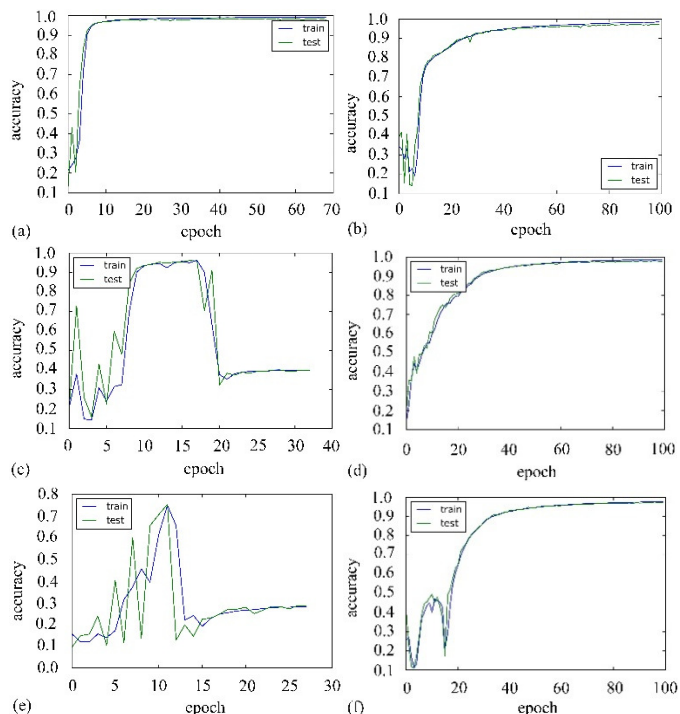

Fig. 1 Accuracies vs. epochs on the train/test datasets obtained by the standard LSTM (top), LSTM1 (middle) and LSTM2 (bottom), with the learning rate coefficients $\eta_0$ = 1e-3 (left) and $\eta_0$ = 1e-4 (right). The difference in epochs is due to the response to the implemented early stopping criterion.

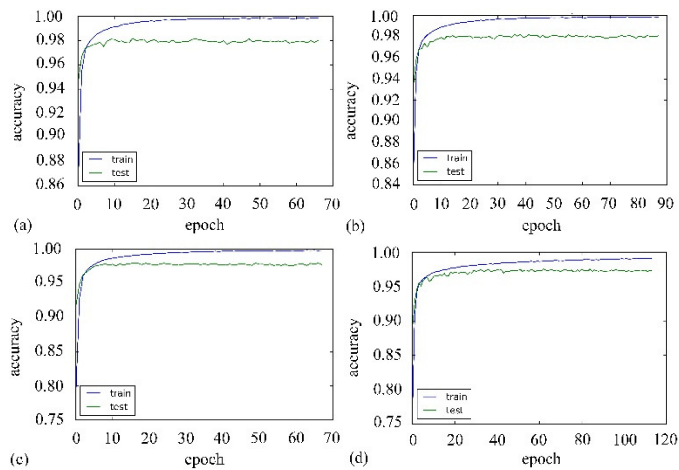

Fig. 3 Accuracies vs. epochs on the train/test dataset obtained by the standard LSTM (a), LSTM1 (b), LSTM2 (c) and LSTM3 (d) with the learning rate coefficient $\eta_0$ = 1e-3. The difference in epochs is due to the response to the early stopping criterion.

From the results of the pixel-wise (long) and row-wise (short) sequence data, it is noted that the three LSTM variants, especially LSTM3, performed closely similar to the standard LSTM in handling the short sequence data.

*B. The IMDB dataset:*

For this dataset, the input sequence from the embedding layer to the LSTM layer is of the intermediate length 128. Table III lists the testing results for various learning coefficients. The standard LSTM and the three variants have show similar accuracies, except that LSTM1 and LSTM2 show slightly lower performance at $\eta_0$=1e-2. Similar to the row-wise MNIST sequence case study, no noticeable fluctuations have been observed for any of the four values of $\eta_0$.

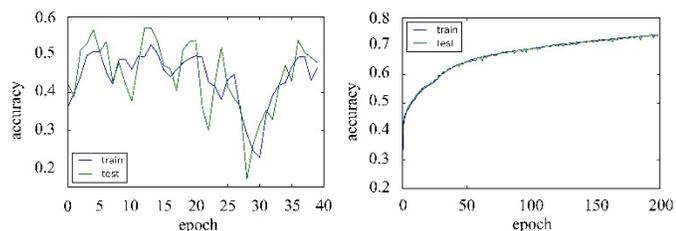

Fig. 2 Accuracies vs. epochs on the train/test datasets obtained by the LSTM3 with the learning rate coefficients $\eta_0$ = 1e-4 (left) and $\eta_0$ = 1e-5 (right). The difference in epochs is due to the response to the early stopping criterion.

Compared to the pixel-wise sequence of length 784, the row-wise sequence form is 28 in length and was much easier (and faster) to train. Table II summarizes the results. All the LSTMs achieved high accuracies at four different $\eta_0$. The standard LSTM, LSTM1 and LSTM2 performed similarly, where they all slightly outperformed the LSTM3. No fluctuation issues were encountered in all the cases. These experiments have used networks with 50 hidden units.

TABLE III
THE BEST ACCURACIES OF DIFFERENT LSTM NETWORKS ON THE TEST SET
AND CORRESPONDING PARAMETER SIZES OF THE LSTM LAYERS

| LSTMs | Learning rate coefficient ($\eta_0$) | | | | # Params |
|---|---|---|---|---|---|
| | 1e-2 | 1e-3 | 1e-4 | 1e-5 | |
| Standard | 0.8467 | 0.8524 | 0.8543 | 0.8552 | 131,584 |
| LSTM1 | 0.7772 | 0.8542 | 0.8532 | 0.8550 | 82,432 |
| LSTM2 | 0.7912 | 0.8512 | 0.8506 | 0.8510 | 82,048 |
| LSTM3 | 0.8279 | 0.8348 | 0.8529 | 0.8548 | 33,280 |

The case for $\eta_0$=1e-5, consistently produced the best results for all the LSTMs and exhibited very similar training/testing profile curves as depicted in Fig. 4.

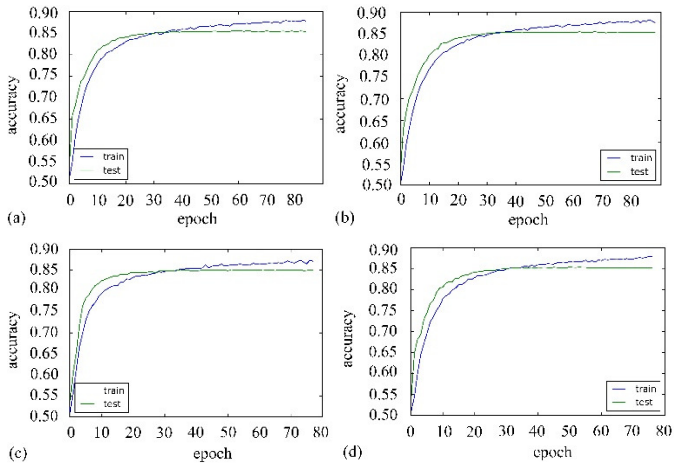

Fig. 4 Accuracies against epochs on the test dataset obtained by the standard LSTM (a), LSTM1 (b), LSTM2 (c) and LSTM3 (d) with the learning rate coefficient $\eta_0$ = 1e-5.

The main benefit of the three LSTM variants is to reduce the number of parameters involved, and thus reduce the computation expense. This has been confirmed from the experiments and as summarized in the three tables above. The LSTM1 and LSTM2 show small difference in the number of parameters and both contain the hidden unit signal in their gates, which explains their similar performance. The LSTM3 has dramatically reduced parameters size since it only uses the bias, an indirectly contained delayed version of the hidden unit signal via the gradient descent update equations. This may explain their relative lagging performance, especially in long sequences. The actual reduction of parameters is dependent on the structure (i.e., dimension) of input sequences and the number of hidden units in the LSTM layer.

## VI. CONCLUSIONS

In this paper, three simplified LSTMs that were defined by eliminating input signal, bias and/or hidden units from their the gate signals in the standard LSTM RNN, were evaluated on the tasks of modeling sequence data of varied lengths. The results confirmed the utility of the three LSTM variants with reduced parameters, which at proper learning rates were capable of achieving the performance comparable to the standard LSTM model. This work represents a preliminary study, and further work is needed to evaluate the three LSTM variants on more extensive datasets of varied sequence length.